\documentclass[10pt,twocolumn,letterpaper]{article}

\usepackage{cvpr}
\usepackage{times}
\usepackage{epsfig}
\usepackage{graphicx}
\usepackage{amsmath}
\usepackage{amssymb}
\usepackage{multirow}
\usepackage{tabu}

\DeclareMathOperator*{\argmax}{argmax}
\newcommand*\pct{\scalebox{0.9}{\%}}

\usepackage[pagebackref=true,breaklinks=true,letterpaper=true,colorlinks,bookmarks=false]{hyperref}

\cvprfinalcopy 

\def\cvprPaperID{3} 

\ifcvprfinal\fi
\begin{document}

\title{Material Segmentation of Multi-View Satellite Imagery}

\author{Matthew Purri\\
Rutgers University\\
New Brunswick, NJ\\
{\tt\small matthew.purri@rutgers.edu}
\and
Jia Xue\\
Rutgers University\\
New Brunswick, NJ\\
{\tt\small jia.xue@rutgers.edu}
\and
Kristin Dana\\
Rutgers University\\
New Brunswick, NJ\\
{\tt\small kdana@ece.rutgers.edu}
\and
Matthew Leotta \\
Kitware \\
Clifton Park, NY \\ {\tt\small matt.leotta@kitware.com}
\and Dan Lipsa \\
Kitware \\
Clifton Park, NY \\ {\tt\small
dan.lipsa@kitware.com}
 \and
Zhixin Li \\
Purdue University \\
West Lafayette, IN  \\
 {\tt\small li2887@purdue.edu}
 \and
Bo Xu \\
Purdue University \\
West Lafayette, IN  \\
 {\tt\small xu1128@purdue.edu}
 \and
Jie Shan \\
Purdue University \\
West Lafayette, IN  \\
 {\tt\small jshan@purdue.edu}}

\maketitle

\begin{abstract}

Material recognition methods use image context and local cues for pixel-wise classification. In many cases only a single image is available to make a material prediction. Image sequences, routinely acquired in applications such as mutli-view stereo, can provide a sampling of the underlying reflectance functions that reveal pixel-level material attributes. We investigate multi-view material segmentation using two datasets generated for building material segmentation and scene material segmentation from the SpaceNet Challenge satellite image dataset~\cite{brown2018large}. In this paper, we explore the impact of multi-angle reflectance information by introducing the \textit{reflectance residual encoding}, which captures both the multi-angle and multispectral information present in our datasets. The residuals are computed by differencing the sparse-sampled reflectance function with a dictionary of pre-defined dense-sampled reflectance functions. Our proposed reflectance residual features improves material segmentation performance when integrated into pixel-wise and semantic segmentation architectures. At test time, predictions from individual segmentations are combined through softmax fusion and refined by building segment voting. We demonstrate robust and accurate pixel-wise segmentation results using the proposed material segmentation pipeline. 

\end{abstract}

\begin{figure}
    \centering
    \includegraphics[width=80mm]{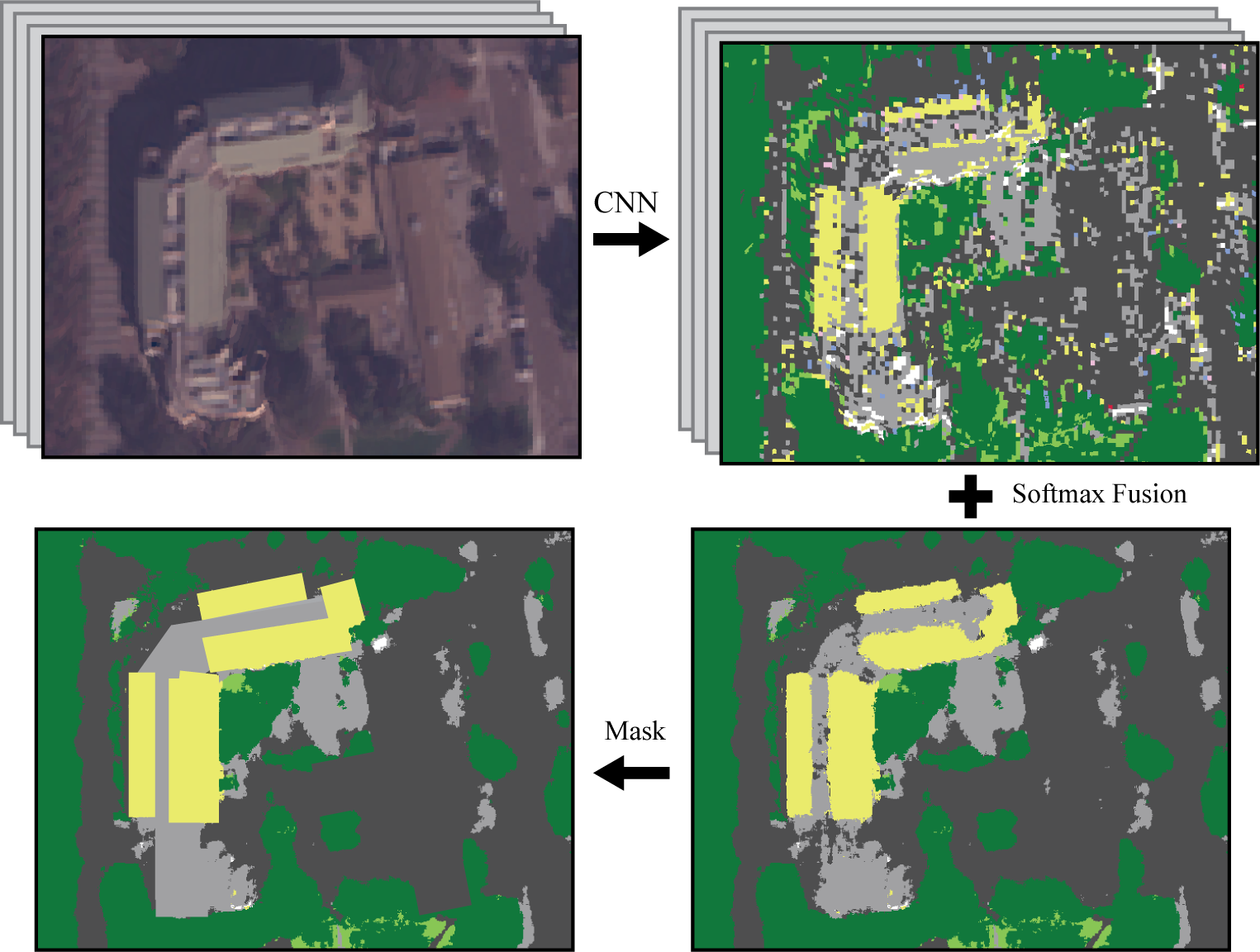}
    \caption{Images from the SpaceNet dataset are orthorectified and individually segmented. The segmented material masks are aggregated through softmax fusion and further refined with a separate building segment mask. }
    \label{fig:post_processing}
\end{figure}

\section{Introduction}
The objective of semantic segmentation is to assign a label to each pixel describing what type of object it belongs to. Similarly for material segmentation, each pixel is assigned a material label. Recognizing materials is important for interacting, understanding, and summarizing complex and novel scenes. Material recognition plays a fundamental role in numerous applications including robotic grasping and pushing~\cite{pinto2017learning, calandra2017feeling, yu2016more}, path navigation for autonomous vehicles~\cite{brandao2016material, ipung2017urban}, quantification of surface albedo for climate modeling~\cite{prado2005measurement}, land-use assessment~\cite{li2013field}, road network recognition~\cite{bajcsy1976computer}, and crop coverage and agricultural assessment~\cite{geipel2014combined}. Material segmentation for satellite imagery is particularly of interest for applications such as road segmentation~\cite{grinias2016mrf, alshehhi2017hierarchical}, land cover albedo analysis~\cite{muster2015spatio}, and tree-cover for fire risk assessment~\cite{forkel2019emergent}. Material segmentation methods rely on texture and reflectance cues while semantic segmentation methods utilize an object's contextual information, shape, and color. For example, doors can have the same shape, color, and contextual information but they could be made up of completely different materials (e.g. wood, metal, plastic). Traditional material recognition techniques measure the reflectance of a surface with a dense sampling of viewing and illumination angles, generating a bidirectional reflectance distribution function (BRDF)~\cite{dana1999reflectance, marschner1999image, lensch2001image}. Gathering a complete BRDF of a surface is infeasible in practice due to the amount of time required for measuring, the need to control the scene illumination, and the ability to access the surface.

\begin{figure*}
    \centering
    \includegraphics[width=170mm, height=70mm]{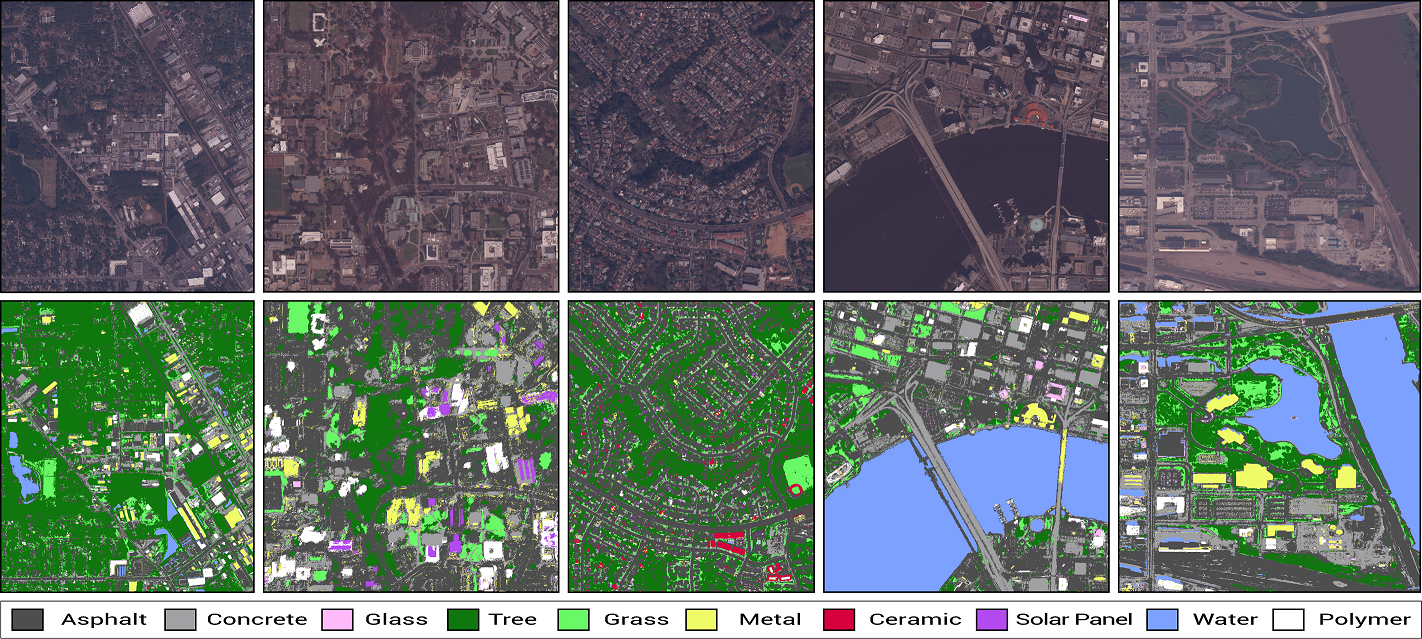}
    \caption{An overview of the tiles and the corresponding ground truth labels used for the scene segmentation dataset. The dense ground truth material masks are generated with a combination of building outline information and a pretrained network. The size of the tiles vary from 9M $pixels^2$ to 81M $pixels^2$.}
    \label{fig:my_label}
\end{figure*}

Large material segmentation datasets such MINC~\cite{bell2015material} contain a single observation per scene of materials. Therefore segmentation algorithms must rely on contextual, texture, and color information instead of multi-angle reflectance information. The NYUv2 RGB-D dataset~\cite{silberman2012indoor} contains 
multiple instances of a scene but the images are labeled are for semantic segmentation instead of material segmentation. Recently, the SpaceNet Challenge dataset~\cite{brown2018large, spacenet}, a multi-instance and multispectral satellite image dataset was made publicly available. In this work, we make use of the images in this dataset for both building material segmentation and scene material segmentation.  The objective of building material segmentation is to assign each pixel belonging to the roof of a building a material label. Building rooftops are constructed from various materials (asphalt, ceramic, glass, etc.) and can contain multiple instances of different materials on a single rooftop. Determining the material makeup of rooftops is useful for building outline extraction, geometry estimation, and realistically rendering 3D building models. We additionally use images from the SpaceNet Challenge dataset for material segmentation of the entire image, a separate and challenging task. 

The major contributions of this work are summarized as follows: 1) we introduce an efficient pipeline for dense material labeling of satellite imagery; 2) the reflectance residual encoding, which combines reflectance measurements from multiple images with non-uniform sampling angles,  improves the material segmentation performance for all tested algorithms; 3) both the softmax fusion and building segment mask post-processing techniques improve material segmentation performance and visual quality.

\section{Related Work}
\label{sec:related_work}
\begin{figure}
    \centering
    \includegraphics[width=80mm]{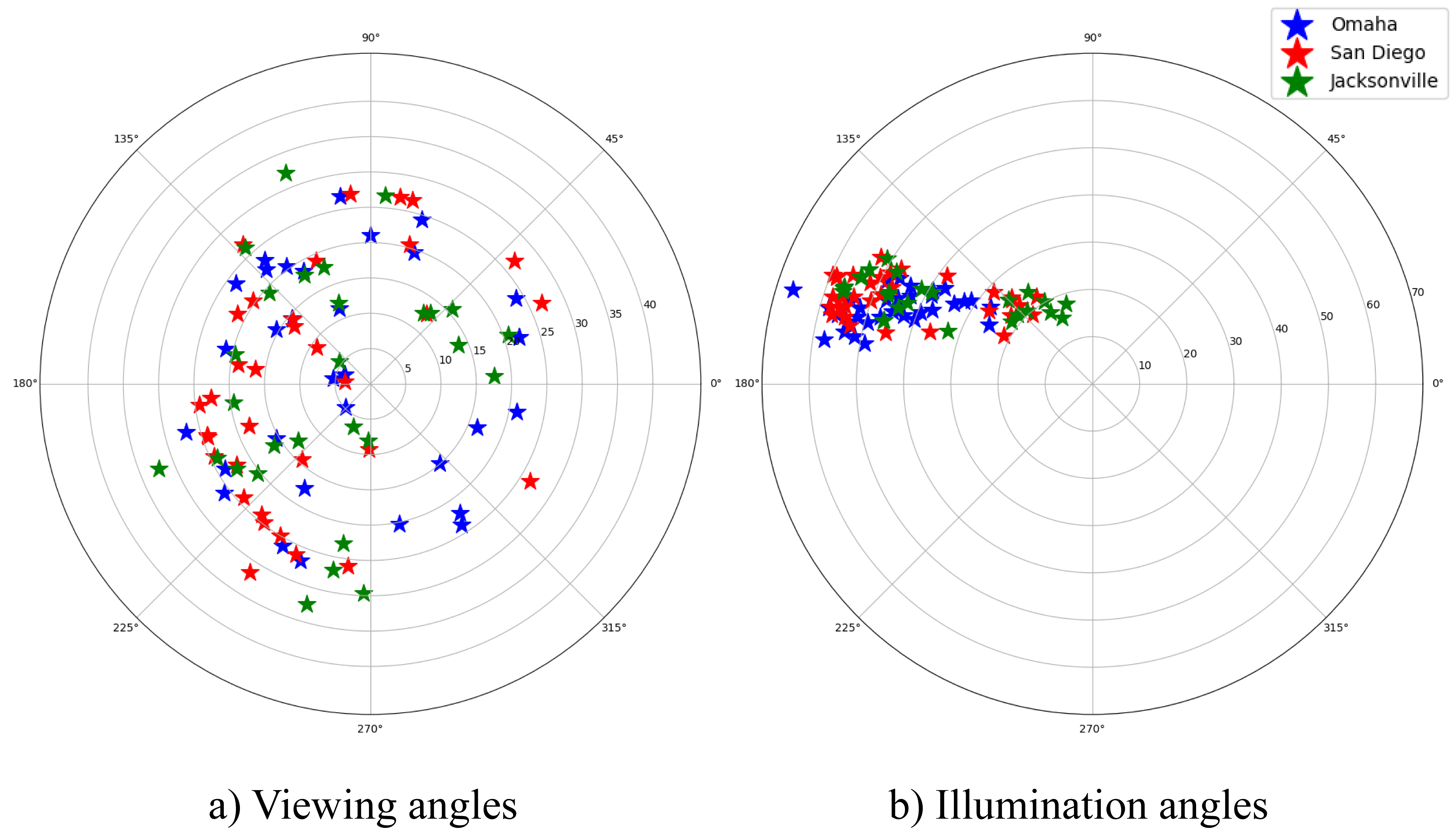}
    \caption{The viewing and illumination sampling angles from the SpaceNet challenge dataset. The viewing angle distribution Omaha (blue), San Diego (red), and Jacksonville (green), shown in a), have different distributions. The distribution of illumination angles, shown in b), vary much less across each region.}
    \label{fig:data_angles}
\end{figure}

\paragraph{Material Recognition}
Prior works in material recognition can be divided into methods that use a single image and methods that use multiple reflectance measurements of a surface. Material recognition from single images rely on texture and reflectance cues to make a reliable prediction~\cite{hu2011toward, Zhang_2017_CVPR, Xue_2017_CVPR}. The resolution of the satellite imagery is not fine enough to discern texture but contains multispectral information useful for material recognition. Zhang et al.~\cite{zhang2016material} show that pixel-wise segmentation from individual hyperspectral images can generate accurate material masks. This approach is similar to our multispectral single angle (MSSA) method which produces a material classification based on the intensity values from a single pixel. Our images contain only 8 wavelengths whereas the images from Zhang et al. contain 28 wavelengths. Methods that rely on reflectance measurements rarely measure an entire BRDF but instead take structured partial samples of a BRDF. For example, reflectance disks~\cite{zhang2016friction}, optimal BRDF sampling~\cite{jehle2010learning, liu2014discriminative}, and BRDF slices~\cite{wang2009material} all provide good material recognition performance with partial reflectance sampling. These methods however require a specialized device or rely on sampling at specific angles. In this work, our dataset is comprised of images taken at non-uniform viewing and illumination angles as shown in Figure~\ref{fig:data_angles}. We naively exploit multi-angle information by concatenating a random selection of images together and perform pixel-wise segmentation. Selecting a subset of the total reflectance measurements limits the representational power of the input features. Building on this work, we introduce a novel reflectance encoding that utilizes all available reflectance measurements which we call the \textit{reflectance residual}. The reflectance residual encoding is inspired by modern dictionary methods for material and texture residual encoding like VLAD~\cite{jegou2012aggregating, arandjelovic2013all}. Unlike prior methods, our dictionary consists of physically measured material BRDFs. The reflectance residual encodes a varying number of input images, numerous wavelengths, and randomly distributed viewing angles into a representational fixed length feature.

\begin{table}[]
    \centering
    \begin{tabu}{c|c|c}
    \tabucline[1.25pt]{-}
        Region & Building & Scene \\ \hline \hline
        San Diego, CA & \checkmark & \checkmark \\
        Jacksonville, FL & \checkmark & \checkmark \\
        Dayton, OH & \checkmark &  \\
        Omaha, NE & & \checkmark \\ \hline
        Ground Truth & Sparse & Dense \\
    \tabucline[1.25pt]{-}
    \end{tabu}
    \caption{An overview of which regions are used for both the building segmentation dataset and the scene material dataset. Additionally, the density of ground truth labeling for measuring performance is described.}
    \label{tab:dataset_info}
\end{table}

\paragraph{Semantic Segmentation}
Segmentation architectures based on the fully convolutional network (FCN)~\cite{long2015fully} achieve state of the art performance on a variety of benchmarks~\cite{zhou2017scene, everingham2010pascal, lin2014microsoft}. The FCN architecture encodes information through a pretrained network which is originally trained on a large classification dataset such as ImageNet~\cite{deng2009imagenet}. The encoded information is then projected back into image space through multiple upsampling layers. The decoding process is unable to recover detailed information lost during downsampling the encoding phase. Methods such as learning upsampling filters through fractionally-strided convolution layers~\cite{noh2015learning, badrinarayanan2017segnet}, replacing convolutional layers with atrous convolutions~\cite{yu2015multi, chen2014semantic, chen2018deeplab}, and the addition of skip connections~\cite{ronneberger2015u, honari2016recombinator, newell2016stacked} have been shown to improve segmentation resolution. Segmentation resolution is of particular importance for satellite material segmentation because the imagery contains a variety of small objects such as air conditioning units, solar panels, and skylights. Inspired by these works, we choose both a FCN with atrous convolutions and a UNet~\cite{ronneberger2015u} with skips connections as our main segmentation architectures. We compare these architectures with a more recent segmentation architecture, EncNet~\cite{zhang2018context}, which achieves state-of-the-art performance on several benchmarks by leveraging global contextual information. It however appears that global contextual information provides limited improvement over the FCN for material segmentation in satellite imagery as shown in Table~\ref{tab:Result_Segmentation}. 

Multi-view semantic segmentation methods utilizing CNNs require each image to be projected into a consistent space. Examples of image space projection include image warping~\cite{ma2017multi} or point cloud generation~\cite{qi2017pointnet, zeng2017multi}. Images used in this work are warped such that the images have pixel consistency, i.e. a pixel coordinate corresponds to the same location in all images. Ma et al.~\cite{ma2017multi} perform image warping to achieve pixel correspondence and then aggregate individual image segmentations through Bayesian fusion and max-pooling of the last feature maps. Inspired by this work, we aggregate multiple individual image segmentations by fusing the outputs of the softmax layer. Multi-image aggregation techniques are found to be useful for smoothing noisy individual segmentations.

\section{Datasets}

The primary datasets used in this work are derivatives of the SpaceNet Challenge dataset~\cite{brown2018large, spacenet}. The SpaceNet Challenge dataset contains both WordView2 and WorldView3 multispectral and panchromatic satellite images from several regions taken over multiple years.  The dataset has been used for challenges involving off-nadir building detection, road network extraction, and building footprint extraction.  The regions of interest in the dataset are medium sized cities and suburbs from the United States.  In this work, only multispectral WorldView3 images are used.  The multispectral images contain eight wavelengths ranging from coastal blue to near infrared red. Images with snow or too much cloud cover are manually removed from the dataset. The dataset is non-uniformly sampled in regard to both the times and the angles the images were taken. 

\begin{figure}
    \centering
    \includegraphics[width=80mm]{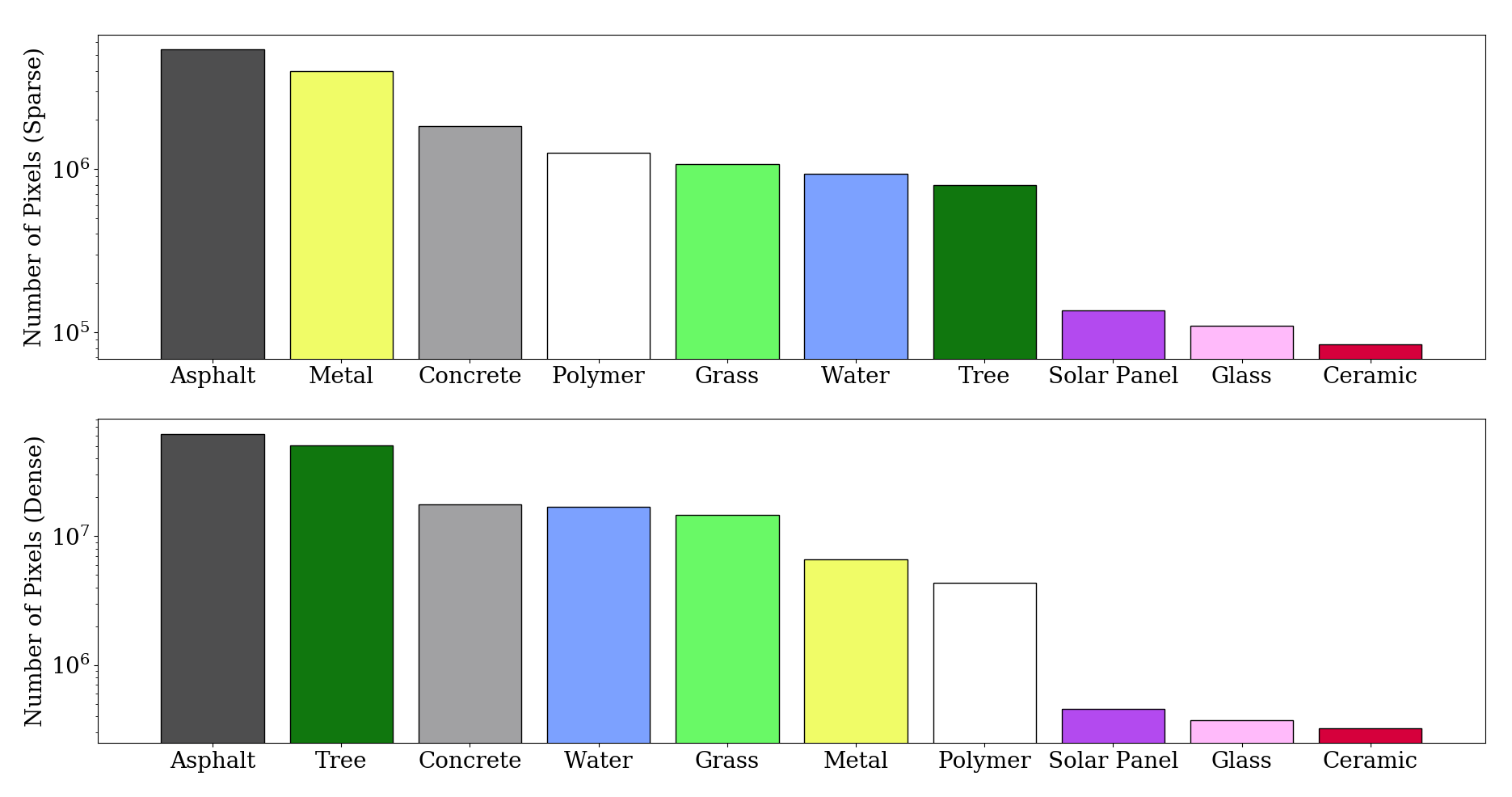}
    \caption{The material distributions for the building segmentation (top) and scene material segmentation (bottom) datasets. }
    \label{fig:class_dist}
\end{figure}

\paragraph{Building Segmentation Dataset}
The objective of building segmentation dataset is to accurately segment the materials of building rooftops in each region. Rooftops can contain a variety of different materials making the task segmentation instead of classification. The regions of interest for this dataset include U.S. cities Jacksonville FL, Dayton OH, and San Diego CA as shown in Table~\ref{tab:dataset_info}. Since the material labels are evaluated only at the location of buildings, the ground truth for this dataset consists of sparse building segment outlines. The dataset comprises 10 different material categories: \textit{asphalt}, \textit{concrete}, \textit{glass}, \textit{tree}, \textit{grass}, \textit{metal}, \textit{ceramic}, \textit{solar panel}, \textit{water}, and \textit{polymer}. For each region a set of tiles are cropped from the original images. The Jacksonville, San Diego, and Dayton areas have two, two, and four tiles respectively, each at different sizes. 

\paragraph{Scene Material Segmentation}
 In contrast to the building segmentation dataset, the goal of the scene material segmentation dataset is to assign a material label to each pixel in the image. The regions contained in this dataset are San Diego, Jacksonville, and Omaha which have two, two, and one tile respectively. This dataset contains the same material classes as the building segmentation dataset. Generated tiles are split into $256 \times 256$ sub-images for input into 2D segmentation algorithms. 

\subsection{Data Processing Pipeline}
The original images found in the SpaceNet Challenge are unwieldy due to their large size. Thus, all images belonging to the same region are first cropped at specified latitude and longitude coordinates. A sparse ground truth material mask is manually created by labeling high confidence regions with material labels.  The ground truth material masks are labeled in a space directly nadir to the ground. In order to correctly assign material labels to off-angle images, a mapping between image space and nadir orientation is required. Images are orthorectified given the image and an elevation model provided by P3D, a module of the Danesfield repository~\cite{Danesfield}.  Images are further aligned using the Lucas-Kanade pixel-wise alignment method. 

WorldView3 images are originally relatively radiometrically calibrated to remove streaks and banding artifacts. The values of each pixel are a function of how much spectral radiance enters the telescope, which is unique to the WorldView3 satellite images. Each channel of the image is  converted to top-of-atmospheric spectral radiance separately by:

\begin{equation}
    L = GAIN \cdot DN \cdot \frac{abscalfactor}{effectivebandwidth} + OFFSET
    \label{eq:Radiometric}
\end{equation}

where the $DN$ corresponds to the raw pixel value, the $GAIN$ and $OFFSET$ are absolute radiometric calibration values, and the $abscalfactor$ is the radiometric calibration factor. The images are further normalized for solar irradiance and sensor radiance by conversation to top-of-atmospheric reflectance by:

\begin{equation}
    R_\lambda = \frac{L_\lambda \cdot d^2 \cdot \pi}{E_\lambda \cdot \cos{\theta_S}}
\end{equation}

Where $L_\lambda$ is the sensor radiance, found in Equation~\ref{eq:Radiometric}, $d$ is the Earth-Sun distance, $E_\lambda$ is the solar irradiance, and $\theta_S$ is the solar zenith angle.  With the images in reflectance units, pixel values can be directly compared to reflectance values measured in material BRDF libraries. 

\subsection{Dense Material Mask Generation}
\label{seg:dense_mask_gen}
Labeling every rooftop in a tile can require thousands of manually generated outlines as well as expert knowledge in material identification from satellite images. This process is difficult to scale and is infeasible for full image material annotation. Instead we develop a semi-automated process that reduces the more tedious aspects of manually labeling to generate fully annotated material masks for each of the tiles in reasonable time frames. A pixel-wise multi-angle convolutional neural network (CNN), further discussed in Section~\ref{sec:1D_alg}, is trained on all manually labeled ground truth data. The trained network evaluates each pixel in the new tile to generate a dense material mask. Generating annotations of dynamic scenes in a shared space inherently leads to label ambiguity. Specific challenges of labeling materials in satellite images from the SpaceNet dataset are seasonal changes, moving objects (e.g. cars), buildings construction, and general outdoor wear and tear of rooftops (e.g. rust or dirt). The resultant dense ground truth material masks are noisy but generally accurate. We employ several noise reducing techniques to improve the ground truth masks used to train our algorithms. Individual image annotation masks are aggregated to produce smoother dense annotations using softmax fusion described in Section~\ref{sec:Post_Processing}. Third party building outlines from U.S. Cities or OpenStreetMap for the tile are gathered according to the coordinates of the tile and projected into image space. For each building outline, an initial material classification is given based on the prediction from the dense mask. An annotator cycles through the building outlines updating any erroneous material classifications and/or adjusting any building outline errors. The time required to label new tiles is significantly reduced through this method. The densely labeled material masks can then be used to train semantic segmentation algorithms. 


\begin{figure*}
    \centering
    \includegraphics[width=160mm]{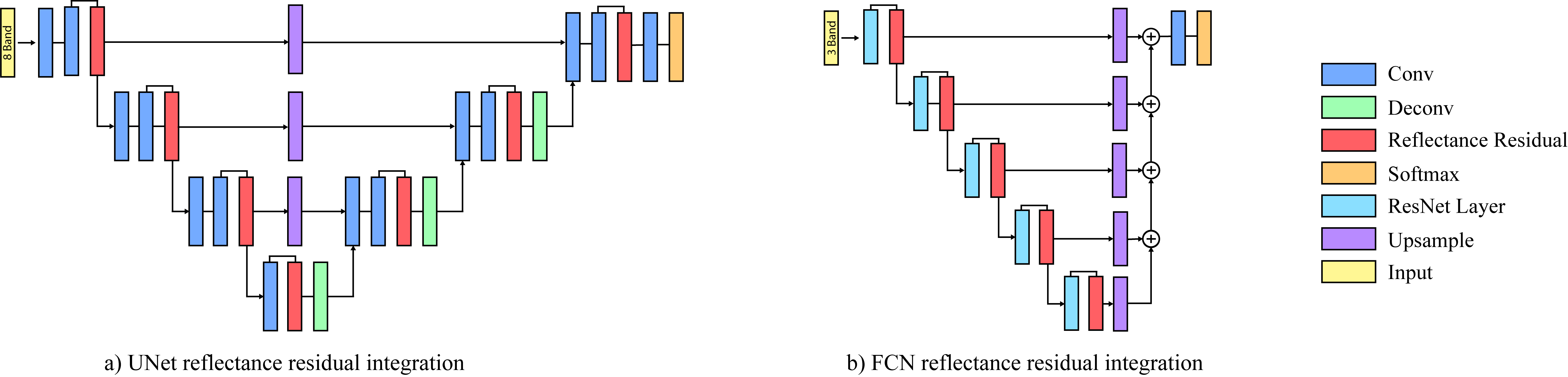}
    \caption{A visualization of how the reflectance residual features are integrated into the UNet and FCN architectures. The resized reflectance residual features (in red) are concatenated to the feature maps at several layers of the architectures.}
    \label{fig:RR+Seg}
\end{figure*}

\section{Algorithms}
We now turn to the task of training deep convolutional neural networks for both building segmentation and material scene segmentation. As described in the Section~\ref{sec:related_work}, CNNs have achieved state-of-the-art performance for semantic segmentation tasks. These networks however require large amounts of near fully annotated ground truth in order to train from scratch and have trouble segmenting small objects reliably. As such, we use 1D networks for the building segmentation dataset and 1D and 2D networks for the material scene segmentation dataset.

\paragraph{Pixel-wise Segmentation}
\label{sec:1D_alg}
The baseline model for pixel-wise prediction is a modified version of an 18 layered Residual Network (ResNet) architecture~\cite{he2016deep}. All 2D convolution and pooling layers from the original structure are replaced with their 1D counterparts. Two models are designed based on this architecture. One method makes predictions based on the raw multispectral pixel information from a single image which we call the multispectral single angle method (MSSA). The input to the MSSA method is an eight length vector, corresponding to the number of channels in the multispectral image. The input vector is upsampled 4x to 32 length before it is used as input.  This method does not take into account the spatial information nor the angular information from the other images. In order to exploit the angular information, we combine pixel intensities from several images into a fixed length vector. Given $N$ images in a region, a fixed number of images $k$ are randomly selected $n \choose k$ and ordered based on their off-nadir viewing angle. The $k$ images are concatenated along the depth axis to create images of size $H \times W \times (8 \cdot k)$. The input to the 1D segmentation algorithm called the multispectral multi-angle method (MSMA) is a $8 \cdot k$ length vector. 

Hyperparameters are shared for training both the MSSA and MSMA methods. Both models are trained for 20 epochs with the Adam weight optimizer~\cite{kingma2014adam}. The learning rate is set to $1e^{-6}$ and is adjusted during training to decrease by a factor of 10 if the training loss plateaus for over 5 epochs. The batch size is set to 128 for training and no data augmentation techniques are used.  The cross entropy objective function is minimized and the class weights for the loss function are set to the inverse frequency of the training set class distribution. The number of images ($k$) used for the MSMA method is set to 15. 


\paragraph{Semantic Segmentation}
\label{sec:exp_seg}
State-of-the-art semantic segmentation architectures employ networks pre-trained on large RGB image classification datasets such as ImageNet. The networks are then fine-tuned on semantic segmentation datasets such as ADE20k~\cite{zhou2017scene} or MSCOCO~\cite{lin2014microsoft} for optimal performance. For the task of material segmentation from satellite imagery both the number of input channels and the type of imagery prevent a majority of the benefits gained from pretrained networks of this kind.  In this work two popular architectures, UNet and FCN, are used for material segmentation on the scene material dataset. The backbone of the FCN architecture is an 18 layer ResNet that is pretrained on the ImageNet dataset. It is then fine-tuned on the satellite imagery. The multispectral satellite imagery is converted to RGB and used as input into the FCN architecture. The UNet architecture is trained from scratch on the full eight channel satellite imagery. 

The semantic segmentation algorithms are trained for 25 epochs with the Adam optimizer.  The learning rate is set to $1e^{-3}$ for all layers except the pretrained layers which have a learning rate of $1e^{-4}$. The learning rate is similarly adjusted as in the pixel-wise algorithm.  The batch size is set to 32 and no data augmentation techniques are used.  The objective function and class weights are the same as in the pixel-wise training. 

 \begin{table}[]
\centering
\resizebox{1.0\linewidth}{!}{
\begin{tabu}{c|c|c|c|c|c|c|c|c}
\tabucline[1.25pt]{-}
\multirow{2}{*}{Method} & \multicolumn{2}{c|}{Dayton} & \multicolumn{2}{c|}{San Diego} & \multicolumn{2}{c}{Jacksonville} & \multicolumn{2}{|c}{} \\ & PixAcc           & mF1
                        & PixAcc           & mF1           & PixAcc          & mF1  & Avg. & Diff.       \\ \hline \hline
MSSA                    &        93.1          &    80.2           &        76.9         &      43.0       & 74.3 & 43.8 & 68.6 & - \\
MSMA                    &           93.2       &    87.5           &        79.9         &      51.2    & 80.7 & 48.9  & 73.6 & 5.0 \\ \tabucline[1.25pt]{-}
\end{tabu}}
\caption{A comparison between the MSSA and MSMA methods with softmax fusion on the building segmentation dataset. The MSMA algorithm consistently outperforms the MSSA method through the utilization of angular information. }
\label{tab:Results_1D_Building_Seg}
\end{table}

\paragraph{Reflectance Residual Encoding}
\label{sec:RR}
As discussed in Section~\ref{sec:1D_alg}, multiple aligned images provide angular information of a scene. As shown in Figure~\ref{fig:data_angles}, the images in each region are taken at much different viewing angles and slightly different illumination angles. The MSMA algorithm makes use of the angular information by randomly sampling images and stacking the pixel values. This method does not encode all of the reflectance information available, instead relying on the CNN to make correlations between the reflectance measurements from different images.  In this work we introduce a novel encoding method that makes use of all available images for a tile by treating each image as a sampling of a BRDF function.  In order to encode the sparse BRDF sampling into a fixed length representation, a library of measured BRDFs are sampled according to the local viewing and illumination angles of a pixel from a set of images. The sampled BRDF values are then compared to the satellite image intensities to create the \textit{reflectance residual} encoding.

Consider a set of $N$ aligned images each with the same number of of pixels. For each pixel $p$, and each image index $j \in [1 \dots N]$ a local viewing and illumination angle ($\theta^v,\phi^v,\theta^i,\phi^i$) is calculated based on the scene surface geometry and the global image viewing and illumination angle. Let $a$ denote this set of angles per pixel sampled by the $N$ images, so that the elements of $a$ are $(\theta_{pj}^v, \phi_{pj}^v, \theta_{pj}^i, \phi_{pj}^i)$.   This set represents the angles in a non-uniform sparse sampling of the underlying BRDF. Let $f$ denote this underlying material BRDF, so that $f(a)$ is the local BRDF sampling. This BRDF sampling is subtracted from  the same sampling for $d$ materials from a dictionary of material BRDFs $M = \{m_1,m_2,...m_d\}$. That is, $m_k(a)$ is subtracted from $f(a)$ for $k\in [1 \dots N]$. The materials in the BRDF database are not required to contain some or any of the target classes and in our case only two materials from the UTIA database \cite{filip14template} are the same as the target classes. The UTIA material BRDF database is used as the BRDF dictionary for this work. The $L2$ norm is computed between the queried intensities from the dictionary BRDF $m_k$ and the  BRDF $f$. For each image the difference is normalized by the intensity of the sampled dictionary element. There is an implicit assumption that the irradiance is uniform  for all image pixels and that all images have been photometrically calibrated. The  residual is calculated as

\begin{equation}
    r_{k,\lambda} = \frac{1}{N} \sum\limits_{p=1}^N \frac{\Vert m_{k,\lambda}(\theta_p^v, \phi_p^v, \theta_p^i, \phi_p^i) - f_{\lambda}(\theta_p^v, \phi_p^v, \theta_p^i, \phi_p^i)\Vert_2^2}{m_{k,\lambda}(\theta_p^v, \phi_p^v, \theta_p^i, \phi_p^i)}
\end{equation}

where $\lambda$ is the wavelength, and $d$ is the user-defined  number of dictionary elements. Each pixel-set generates an $N \times d \times \lambda$ tensor after BRDF dictionary comparison. The mean of the values over $N$ images generates a $d \times \lambda$ encoding matrix which we call the reflectance residual. The reflectance residual is a fixed length representation of the comparison between the image-set sampled BRDF and the dictionary of material BRDF. The reflectance residual features are generated for each pixel and can be generated in parallel to form a reflectance residual over a set of images.

The pixel-wise reflectance residual (RR) encoding is used as input to the 1D algorithm and called the \textit{per-pixel RR} method. The reflectance residual encoding is integrated into the semantic segmentation features by resizing and concatenating the RR encoding to intermediate feature maps in the network. The addition of the reflectance residual encoding into both the FCN and UNet architectures is shown in Figure~\ref{fig:RR+Seg}. The training procedure for the pixel-wise method or segmentation methods are unchanged with the addition of the \textit{reflectance residual}.

\begin{table}[]
\centering
\resizebox{0.97\linewidth}{!}{
\begin{tabu}{c|c|c|c|c|c|c}
\tabucline[1.25pt]{-}
Fusion & Voting & Dayton & San Diego & Jacksonville & Avg. & Diff.   \\\hline \hline
 &     & 78.2 & 67.5 & 70.2 & 72.0 & - \\
\checkmark &  & 93.2 & 79.9 & 80.7 & 84.6 & 12.6 \\
\checkmark & \checkmark & 97.3 & 80.3 & 84.0 & 87.2 & 2.6 \\
\tabucline[1.25pt]{-}
\end{tabu}}
\caption{Ablation study of post-processing techniques to refine instance-wise material segmentations. Both softmax fusion and building segment voting improve the average pixel-wise accuracy of the MSMA method.}
\label{tab:Result_Post_Ablation}
\end{table}

\begin{table*}[]
\centering
\begin{tabu}{c|c|c|c|c|c|c|c|c|c|c|cc}
\tabucline[1.25pt]{-}
\hline
\multirow{2}{*}{Dim.} & \multirow{2}{*}{Model Name} & \multicolumn{3}{c|}{Jacksonville} & \multicolumn{3}{c|}{San Diego} & \multicolumn{3}{c|}{Omaha} &                           &       \\
                      &                             & PixAcc      & mF1      & mIoU     & PixAcc     & mF1     & mIoU    & PixAcc   & mF1    & mIoU   & \multicolumn{1}{c|}{Avg.} & Diff. \\ \hline \hline
\multirow{5}{*}{2D}   & FCN                         & 66.0        & 27.1     & 19.7     & 71.0       & 25.7    & 18.9    & 69.4     & 37.4   & 29.8   & \multicolumn{1}{c|}{40.6} & -     \\
                      & EncNet~\cite{zhang2018context}              & 66.1        & 27.3     & 20.5     & 71.2       & 25.4    & 18.8    & 69.6     & 37.8   & 29.9   & \multicolumn{1}{c|}{40.7} & 0.1   \\
                      & FCN + RR                    & 69.6        & 27.9     & 19.9     & 73.8       & 26.3    & 19.5    & 70.1     & 37.6   & 30.1   & \multicolumn{1}{c|}{41.6} & 1.0   \\ \cline{2-13}
                      & UNet                        & 80.1        & 42.1     & 33.1     & 75.7       & 26.5    & 21.1    & 70.5     & 35.9   & 27.9   & \multicolumn{1}{c|}{45.8} & -     \\       
                      & UNet + RR                   & 80.1        & 46.1     & 35.3     & 75.2       & 31.9    & 23.5    & 71.9     & 37.4   & 29.5   & \multicolumn{1}{c|}{47.9} & 2.1   \\ \hline
\multirow{2}{*}{1D}   & MSSA         & 47.4        & 15.2     & 11.1     & 64.6       & 36.4    & 31.5    & 51.6     & 26.9   & 18.7   & \multicolumn{1}{c|}{33.7} & -     \\
                      & Per-Pixel RR                & 80.8        & 39.9     & 31.0     & 80.6       & 41.9    & 32.6    & 65.2     & 37.0   & 27.3   & \multicolumn{1}{c|}{48.5} & 14.8  \\ \tabucline[1.25pt]{-}
\end{tabu}%
\caption{The performance of the pixel-wise and image segmentation algorithms on the scene material segmentation dataset. We observe a consistent score improvement with the addition of the reflectance residual (RR) features. The per-pixel RR method on average outperforms the pixel-wise multispectral multi-angle (MSSA) algorithm and the image segmentation algorithms.}
\label{tab:Result_Segmentation}
\end{table*} 

\paragraph{Post Processing}
\label{sec:Post_Processing}
The pixel-wise MSSA and image segmentation algorithms generate material predictions image-wise while the MSMA method samples a fixed number of images to generate a single predication. The MSMA method can be resampled to generate another prediction. Softmax fusion aggregates the instance-wise predictions into a single prediction by adding the outputs of individual softmax predictions. Consider the cross entropy output distribution for a pixel $z^i$ corresponding to an image $i$. The length of vector $z_i$ is $C$, the number of material classes.  The combined prediction is computed by

\begin{equation}
     y = \argmax_c \sum_i \frac{e^{z^i}}{\sum_c^C e^{z_c^i}}
\end{equation}
where $y$ is the aggregated prediction. 

In addition to softmax fusion aggregation, a local voting technique that utilizes building segment masks is applied to the resultant prediction. A separate module of the Danesfield repository generates a building segment mask for dividing complex building geometries into many primitive shapes~\cite{Danesfield}.  We employ this mask after the softmax fusion to cluster pixels belonging to the same building segment so that noisy predictions are reduced as shown in Figure~\ref{fig:post_processing}. For each building segment the most common  material class is assigned to all pixels in that segment. We call this post processing technique \textit{building segment voting}. 

\section{Results}

\paragraph{Building Segmentation}
Table~\ref{tab:Results_1D_Building_Seg} shows the results of the MSSA and the MSMA algorithms on the building segment dataset. In every region the MSMA method outperforms the MSSA method, highlighting the importance of angular information for material classification. Both algorithms are trained on all regions excluding the evaluation region and the results are aggregated using softmax fusion method. Not only does the MSMA method outperform the MSSA method in total pixels correct but it also performs better across a majority of the material classes according to the mean F1-scores. This suggests that the MSMA method is not assigning the most likely material but is able to identify less common materials such as glass, solar panel, and ceramic.

\paragraph{Post Processing Ablation Study}
We find that aggregating the results from other instances significantly improves the segmentation performance as shown in Table~\ref{tab:Result_Post_Ablation}. The baseline method in Table~\ref{tab:Result_Post_Ablation} is the average pixel accuracy for the MSMA method resampled 10 times for a given region. The segmentation performance improves by 12.6\% when aggregating the 10 predictions with softmax fusion. We find similar but less dramatic improvements for the MSSA method. We conclude that at some angles, materials are difficult to determine and combining several image predictions leads to smoother and more accurate results. The results from softmax fusion are further improved from the building segment voting method.  The voting process removes some of the warping prediction noise from orthorectification not removed with softmax fusion, see Figure~\ref{fig:post_processing}.

\begin{figure*}
    \centering
    \includegraphics[width=170mm, height=95mm]{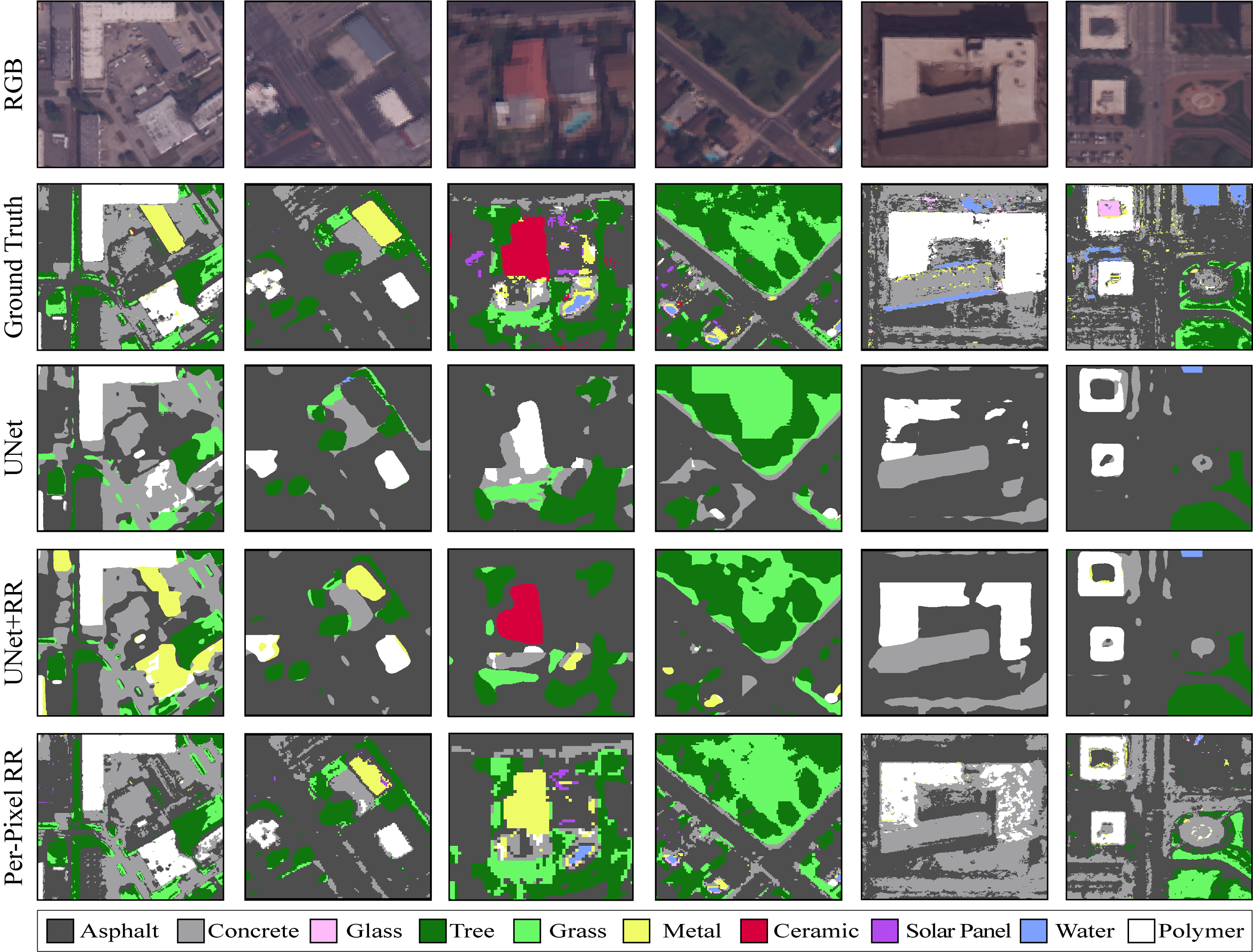}
    \caption{A comparison of segmentation performance on the scene material segmentation dataset. Observe that the UNet network does not correctly classify metal buildings and sections of buildings are not labeled consistently. The UNet model with reflectance residual features is able to determine metal and ceramic buildings as well as generate a consistently label sections of the buildings. The pixel-wise reflectance residual method gives segmentations with sharp borders and does better than the other methods on the vegetation classes. The {\it per-pixel RR} method is the only one able to identify small swimming pools in columns 3 and 4. The dense ground truth labels are generated by a pixel-wise network trained on sparse labels (25\% of total pixels labeled) and are not perfect.}
    \label{fig:Results_Qual_Seg}
\end{figure*}

\paragraph{Pixel-wise Segmentation}
We evaluate the performance of both the MSSA and per-pixel RR methods on the material segmentation dataset. The results shown in Table \ref{tab:Result_Segmentation} correspond to softmax fusion over all images for the tile. The MSSA method, which does not utilize the reflectance residual features, performs worse compared to the {\it per-pixel RR} method. A single view of a material without spatial information is not enough to reliably predict materials from the material segmentation dataset. The reflectance residual features are able to encode both multispectral and multi-angle information, which leads to performance improvements on average of 21.0\pct/13.4\pct/9.9\pct~compared to only using multispectral information. The per-pixel methods perform relatively lower on the Omaha region and is likely a result of not having an Omaha tile in the training set. Differences in atmospheric conditions of the scene affect the performance of both 1D segmentation methods. Due to the lack of training images from the Omaha region, the 1D segmentation results are unable to generalize as well do to the unseen atmospheric noise. 

\paragraph{2D Segmentation}
Table~\ref{tab:Result_Segmentation} compares the performance of both the UNet and FCN architectures with and without the integration of reflectance residual features. Additionally, EncNet a state-of-the-art architecture derived from FCN is compared to gauge the relative improvement of the reflectance residual encoding. EncNet narrowly surpasses the performance of the FCN network for the scene material segmentation  dataset. The addition of reflectance residual features improves the performance of FCN 10x more relative to EncNet improvement. On average the UNet architecture outperforms the FCN network in all of the metrics. The UNet architecture has been shown to perform 
well when the size of the dataset is small \cite{ronneberger2015u, milletari2016v}. In our case the SpaceNet dataset is smaller than other segmentation datasets such as Pascal VOC or ADE20K. The integration of reflectance residual features into each network improves the performance across all regions. The addition of reflectance residuals improves the UNet architecture performance by 0.3\pct/3.7\pct/2.1\pct~in terms of pixel-wise accuracy, average F1 score, and mean IoU. Qualitatively the segmentation results of the UNet with reflectance residuals outperforms the UNet without reflectance residual on more difficult classes such as metal and ceramic as shown columns 1 and 3 in Figure \ref{fig:Results_Qual_Seg}. Across Figure \ref{fig:Results_Qual_Seg}, the UNet architecture with reflectance residuals generates more accurate boundaries and better identifies less common materials. 

\paragraph{1D vs 2D segmentation}
The performance of the 2D semantic segmentation networks are compared with the 1D pixel-wise segmentation networks quantitatively in Table \ref{tab:Result_Segmentation} and qualitatively in Figure \ref{fig:Results_Qual_Seg}. The UNet+RR network achieves the highest pixel-wise accuracy while the 1D {\it per-pixel RR} network outperforms all other methods on average F1 and mean IoU scores. For each region the highest metrics come from networks that utilize the reflectance residual features. From Figure \ref{fig:Results_Qual_Seg}, we can visually see how the reflectance residual features improve the baseline UNet method. The baseline UNet architecture is unable to identify materials such as ceramic, metal, or water reliably. Results from the first, second and third column show that those difficult materials are more likely to be predicted correctly with networks utilizing reflectance residual features. The reflectance residual features also improve the shape of the predictions as seen in columns 3 and 5 from Figure \ref{fig:Results_Qual_Seg}. 1D methods lack the spatial information utilized by 2D methods but appear to make up for it in prediction resolution. The segmentation networks are unable to correctly identify water in small pools as shown in columns 3 and 4 of Figure \ref{fig:Results_Qual_Seg} while the {\it per-pixel RR} network consistently identifies them. From columns 4 and 6 we see that unintuitively the {\it per-pixel RR} method better distinguishes between the tree and grass vegetation classes.  

\section{Conclusion}
We have introduced several novel methods for material segmentation for multi-view satellite imagery and compared them to state-of-the-art semantic segmentation architectures.  We show that angular information is an important cue for material segmentation.  The utilization of angular information improves the performance of 1D and 2D algorithms on both datasets. Specifically a physically based encoding method, reflectance residual, is introduced and integrated into semantic segmentation networks for increased performance. Additionally an efficient method for generating accurate fully annotated material masks for satellite imagery is provided. We use prediction aggregation and building segment masks to improve the segmentation results on both datasets.

\section*{Acknowledgements}
Supported by the Intelligence Advanced Research Projects Activity (IARPA) via
Department of Interior/ Interior Business Center (DOI/IBC) contract number
D17PC00286. The U.S. Government is authorized to reproduce and distribute reprints
for Governmental purposes notwithstanding any copyright annotation thereon.
Disclaimer: The views and conclusions contained herein are those of the authors and
should not be interpreted as necessarily representing the official policies or
endorsements, either expressed or implied, of IARPA, DOI/IBC, or the U.S.
Government.

{\small
\bibliographystyle{ieee}
\bibliography{egbib}
}

\end{document}